\documentclass[runningheads]{llncs}
\usepackage[T1]{fontenc}
\usepackage{graphicx,verbatim}
\usepackage{caption}
\usepackage{subcaption}
\usepackage{booktabs}
\usepackage{tabularx}
\usepackage{multirow}

\begin{document}
\title{Deep Learning-based Alignment Measurement in Knee Radiographs}
\author{%
Zhisen Hu$^{*}$\inst{1} \and
Dominic Cullen\inst{1,2} \and
Peter Thompson\inst{1} \and
David Johnson\inst{3,4,5} \and
Chang Bian\inst{1} \and
Aleksei Tiulpin\inst{6} \and
Timothy Cootes\inst{1} \and
Claudia Lindner\inst{1}%
}
\authorrunning{Z. Hu et al.}
\institute{Division of Informatics, Imaging and Data Sciences, The University of Manchester, Manchester M13 9PT, United Kingdom \and Northern Care Alliance NHS Foundation Trust, Salford M6 8HD, United Kingdom \and Department of Trauma and Orthopaedics, Stockport NHS Foundation Trust, Stepping Hill Hospital, Stockport SK2 7JE, United Kingdom \and School of Health and Society, University of Salford, Salford M6 6PU, United Kingdom \and School of Biological Sciences, The University of Manchester, Manchester M13 9PL, United Kingdom \and Research Unit of Health Sciences and Technology, University of Oulu, Oulu 90220, Finland \\
\email{zhisen.hu@postgrad.manchester.ac.uk}}

\maketitle              % typeset the header of the contribution

\begingroup
\renewcommand{\thefootnote}{}  % remove footnote number
\footnotetext[1]{*Corresponding Author}
\addtocounter{footnote}{-1}    % prevent counter from increasing
\endgroup

\begin{abstract}
Radiographic knee alignment (KA) measurement is important for predicting joint health and surgical outcomes after total knee replacement. Traditional methods for KA measurements are manual, time-consuming and require long-leg radiographs. This study proposes a deep learning-based method to measure KA in anteroposterior knee radiographs via automatically localized knee anatomical landmarks. Our method builds on hourglass networks and incorporates an attention gate structure to enhance robustness and focus on key anatomical features. To our knowledge, this is the first deep learning-based method to localize over 100 knee anatomical landmarks to fully outline the knee shape while integrating KA measurements on both pre-operative and post-operative images. It provides highly accurate and reliable anatomical varus/valgus KA measurements using the anatomical tibiofemoral angle, achieving mean absolute differences $\mathord{\sim}1$° when compared to clinical ground truth measurements. Agreement between automated and clinical measurements was excellent pre-operatively (intra-class correlation coefficient (ICC) = 0.97) and good post-operatively (ICC = 0.86). Our findings demonstrate that KA assessment can be automated with high accuracy, creating opportunities for digitally enhanced clinical workflows.

\keywords{Knee alignment \and Landmark localization \and Deep learning \and Hourglass \and Anatomical tibiofemoral angle}
% Authors must provide keywords and are not allowed to remove this Keyword section.

\end{abstract}
\section{Introduction}
Knee osteoarthritis (OA) is a common and significant health issue that heavily burdens healthcare systems~\cite{jin2020incidence}. Total knee replacement (TKR) may be offered as treatment for end-stage knee OA. Nevertheless, TKR is invasive involving prosthesis implantation at the knee joint, and around 10$\%$ of patients are dissatisfied following TKR~\cite{ozden2025what,defrance2023are}. Pre-operative and post-operative knee alignment (KA) affects the outcomes following TKR, with radiographs revealing anomalies such as deformities of the femur and tibia, as well as incorrect positioning of the implants~\cite{ritter2011effect,ritter2013preoperative}. Accurate assessment of KA in radiographs is important for successful treatment outcomes and long-term joint health. Traditional KA measurement methods are manual, time-consuming, and require long-leg radiographs. However, long-leg radiographs are not always undertaken in clinical practice, and standard anteroposterior (AP) knee radiographs are often the main imaging modality. Automated methods for measuring KA in AP knee radiographs are potentially clinically valuable for reducing the cost and improving the efficiency of the knee OA treatment pathway.

\begin{figure}[ht]
    \centering
    \subcaptionbox{\label{fig:Preop_ap_eg}}{
    \includegraphics[width=0.25\textwidth]{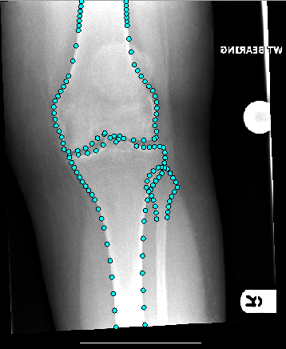}
    }
    % \hfill
    \subcaptionbox{\label{fig:Postop_ap_eg}}{
    \includegraphics[width=0.215\textwidth]{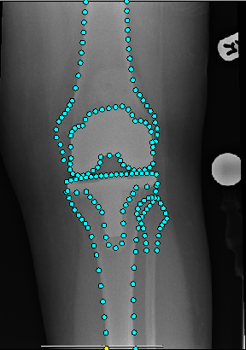}
    }
    % \hfill
    \subcaptionbox{\label{fig:aTFA}}{
    \includegraphics[width=0.17\textwidth]{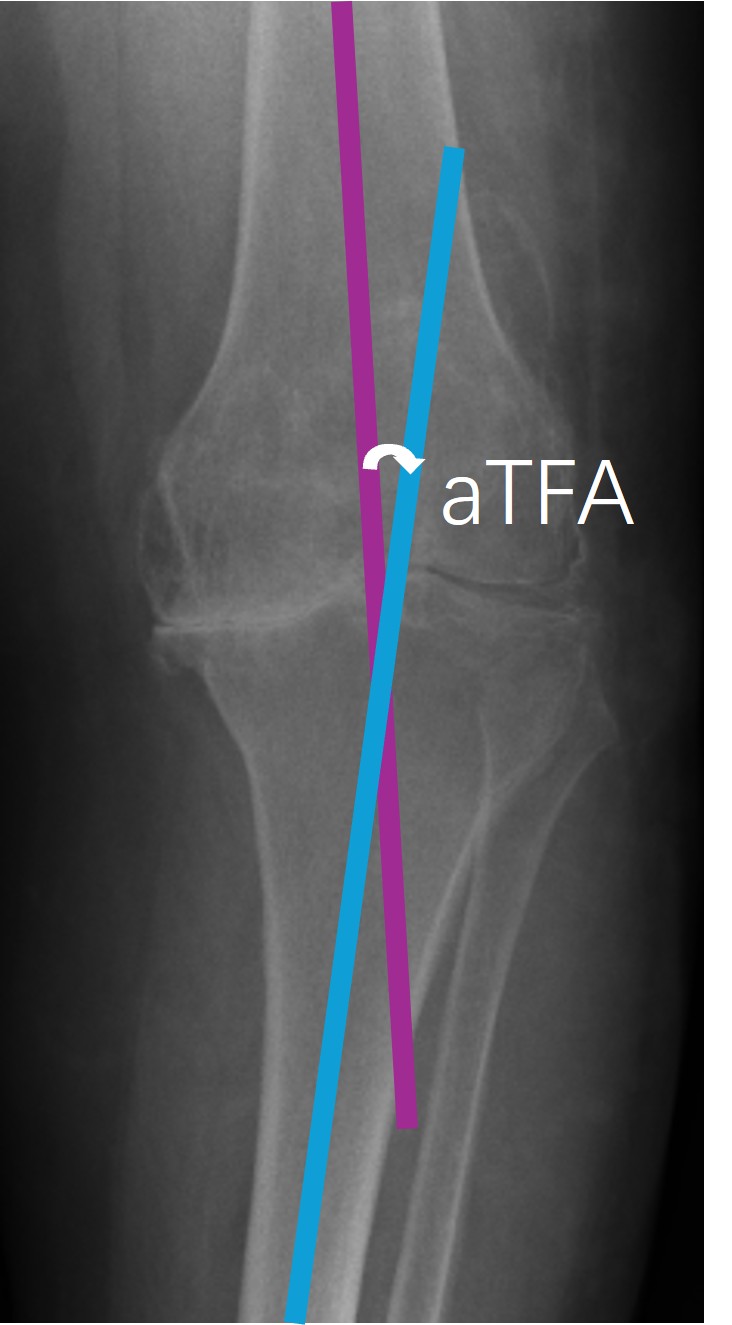}
    }
\caption{An illustration of AP knee radiographs with corresponding anatomical landmarks in (a) pre-operative and (b) post-operative images, and (c) the anatomical tibiofemoral angle (aTFA) measurement. aTFA is defined by the angle between the anatomical femoral axis (purple line) and tibial axis (blue line).}
\label{fig:kneeexample}
\end{figure}

Knee anatomical landmark positions (Fig.~\ref{fig:Preop_ap_eg} and~\ref{fig:Postop_ap_eg}) are often used for automatically generating KA measurements~\cite{ye2020development,nam2023key}. Recently, machine learning and deep learning have been widely used for localizing knee anatomical landmarks in radiographs. One of the state-of-the-art methods of knee landmark localization is based on a combination of random forest regression voting (RFRV) with constrained local model (CLM) ﬁtting~\cite{lindner2014robust,lindner2013accurate}. State-of-the-art deep learning-based methods include the study by Tiulpin et al.~\cite{tiulpin2019kneel}, which used hourglass networks~\cite{newell2016stacked} to regress the knee landmark positions from AP knee radiographs. Several other methods used U-Nets~\cite{ronneberger2015unet} to localize pelvis and hand landmarks~\cite{davison2019landmark,payer2019integrating}. Our method is based on the hourglass network architecture in~\cite{tiulpin2019kneel} and combines the network with an attention gate (AG) structure~\cite{oktay2018attention} to better focus on target joint shapes in knee radiographs.

This study proposes a deep learning-based approach to automatically localize knee anatomical landmarks and measure varus and valgus KA using the anatomical tibiofemoral angle (aTFA). In both pre-operative and post-operative AP knee radiographs, the aTFA is defined by the angle between the anatomical femoral and tibial axes (Fig.~\ref{fig:aTFA}). To our knowledge, this is the first deep learning-based study to localize over 100 anatomical landmarks in knee radiographs and integrate KA measurements on both pre-operative and post-operative images. 

\noindent \textbf{Contributions:}

\noindent 1) We compare our method to~\cite{cullen2025an}, demonstrating better accuracy in landmark localization and improved performance in KA measurements.

\noindent 2) We demonstrate how different methods of generating KA measurements affect the agreement with ground truth measurements.

\section{Method}

\begin{figure}[ht]
    \centering
    \includegraphics[width=0.95\textwidth]{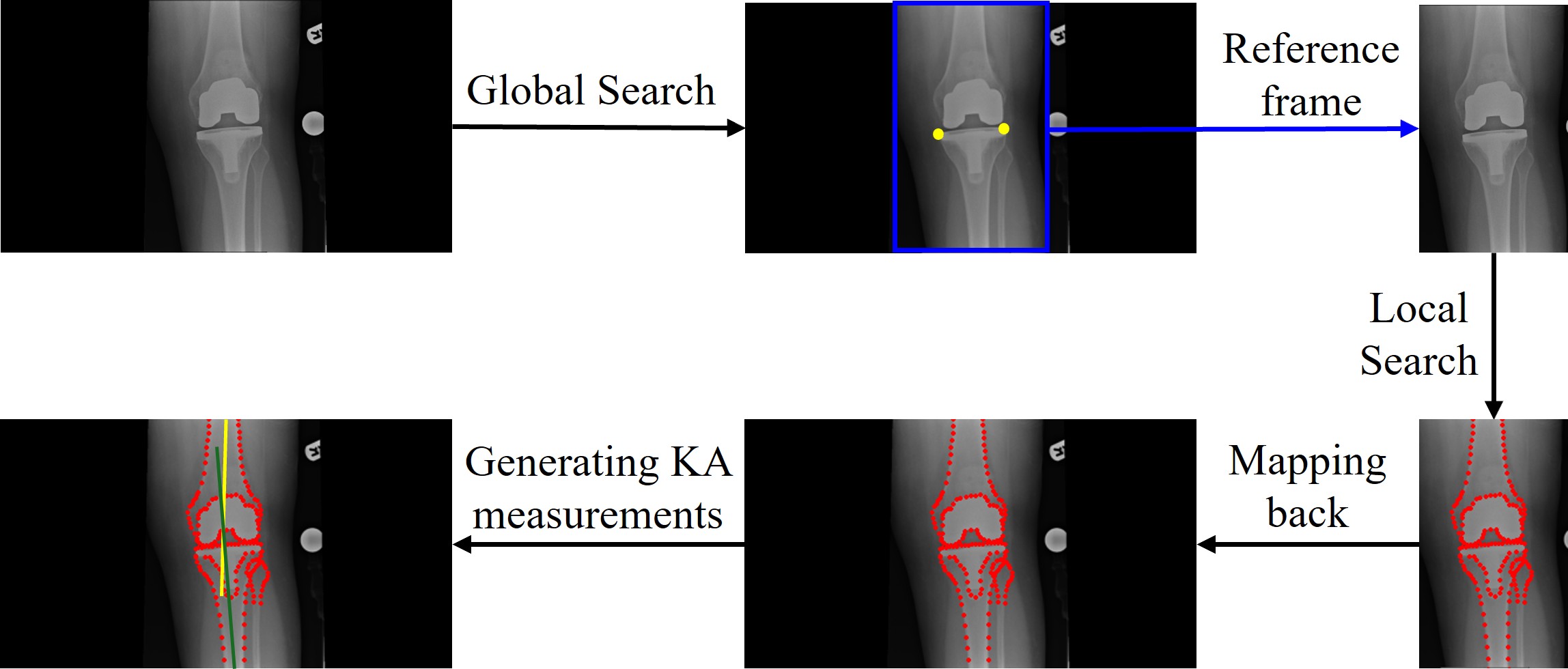}
    \caption{The workflow of our automated KA measurement approach. The global search model searches across the entire image and initializes two reference points (yellow points), which establish the approximate position, orientation, and scale of a reference frame (region of interest). Then the local search model localizes over 100 knee landmarks (red points) within the reference frame to outline the shape of the knee joint. The landmarks are mapped back to the original image for comparison with original manual annotations. The KA measurements are then generated from these landmark positions.}
    \label{fig:workflow}
\end{figure}

The workflow of our automated KA measurement approach is shown in Fig.~\ref{fig:workflow}. The knee landmarks are localized first, and subsequently the KA measurements are generated based on the landmark positions.

\subsection{Data}
Our dataset is composed of anonymized standard AP knee radiographs from TKR patients. To simplify the analysis, all right knee radiographs were flipped horizontally to appear as left knee radiographs. All radiographs were retrospectively collected from Stockport NHS Foundation Trust (approved by the Health Research Authority, IRAS 244130). All subjects underwent primary TKR and had no revision surgery within three years after TKR. Our dataset consists of 566 pre-operative and 457 one-year post-operative images for training, and 376 patient image pairs (pre-operative and one-year post-operative) for testing~\cite{cullen2025an}. Landmarks were defined along the distal femur and proximal tibia/fibula to capture the knee joint, including implants in the post-operative images (see Fig.~\ref{fig:Preop_ap_eg} and~\ref{fig:Postop_ap_eg}). The pre-operative and post-operative images were manually annotated with 134 and 181 landmarks, respectively.  

\subsection{Landmark Localization}
\begin{figure}[ht]
    \centering
    \includegraphics[width=0.95\textwidth]{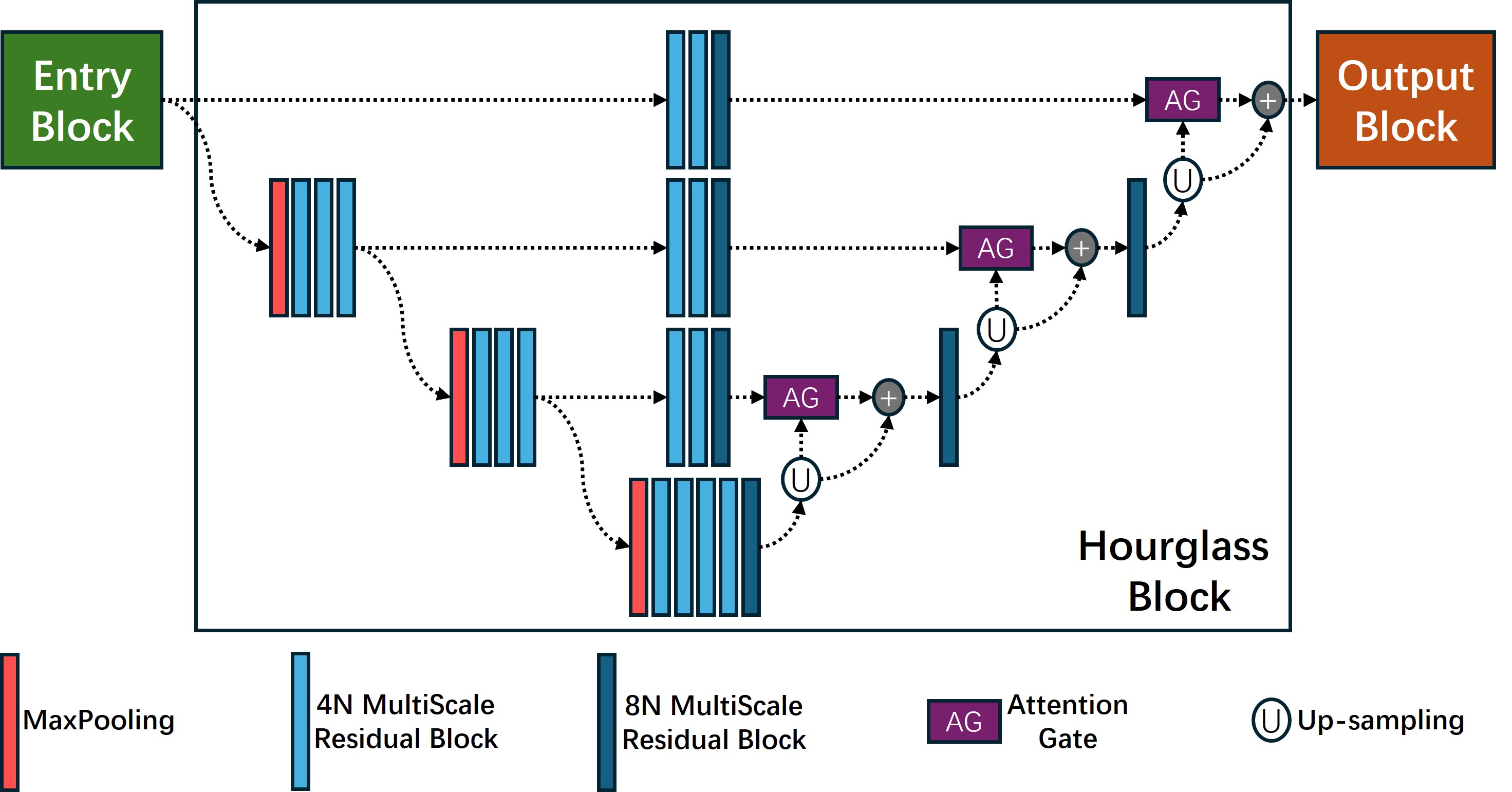}
    \caption{Model architecture with an hourglass network of depth d=4 combined with AGs. Here, N is the width (initial number of channels) of the network.}
    \label{fig:network}
\end{figure}

A deep learning-based system using hourglass networks was trained to localize the anatomical knee landmarks. Our network structure (shown in Fig.~\ref{fig:network}) is similar to the hourglass network in~\cite{tiulpin2019kneel}. AG blocks similar to~\cite{oktay2018attention} are used to filter the features passed from the upper-level blocks of the hourglass network. Our automated landmark localization system consists of two stages: global search and local search (with an independent hourglass network for each stage).

The global search aims to narrow down the search area for the subsequent local search stage. We use a 4-layer hourglass network to scan the entire knee radiograph to identify two reference points on the knee joint. In our case, we chose two landmarks of the local search model for the purpose of initialization. The two reference points have a fixed position in the reference frame, which is centered around the region of interest identified in the global search. The position, orientation, and scale of the reference frame are defined by the two reference points. 

The local model searches in the more confined reference frame. The objective is to accurately localize specific landmarks on the target object. We use a 6-layer hourglass network to localize the knee landmarks. The landmark positions are then mapped back to the original image using the position, orientation, and scale obtained from the global search. 

\subsection{Knee Alignment}

We measured varus/valgus KA in standard AP knee radiographs both pre-operatively and post-operatively using the aTFA (Fig.~\ref{fig:aTFA}). Varus and valgus were defined as negative and positive deviations from zero, respectively. We included two sets of point-based measurements in our experiments. \textbf{Automated measurements} were assessed based on a subset of the \textit{automatically localized} pre-operative and post-operative landmark positions in the 376 test patients. \textbf{Manual measurements} were generated based on a subset of the \textit{manually annotated} landmark positions in pre-operative and post-operative images of the 376 test patients and were used as the manual ground-truth.

In addition, we also included a set of \textbf{clinical measurements} which were directly measured in a clinical setting with a Picture Archiving and Communication System (PACS)-integrated measurement facility by an orthopedic surgeon. Clinical measurements were obtained for a random subset of 50 test patients from the 376 test patients. Two clinical measurements were taken for each image, with a 7–10 day interval between them, and the second measurement was made without knowledge of the first. The mean of the two measurements was used as the clinical ground truth.

\begin{figure}[ht]
    \centering
    \subcaptionbox{\label{fig:Preop_KA}}{
    \includegraphics[width=0.30\textwidth]{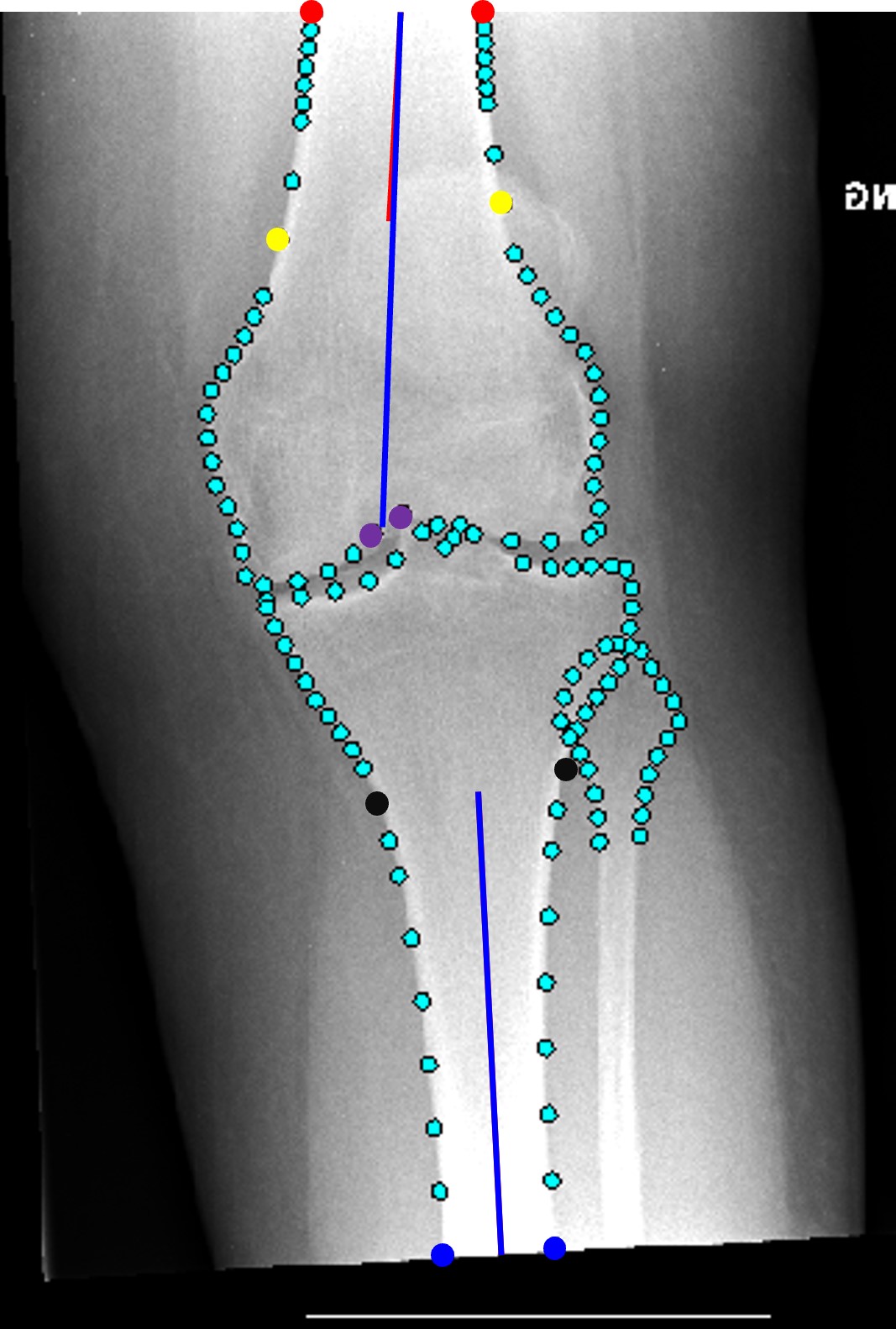}
    }
    % \hfill
    \subcaptionbox{\label{fig:Postop_KA}}{
    \includegraphics[width=0.287\textwidth]{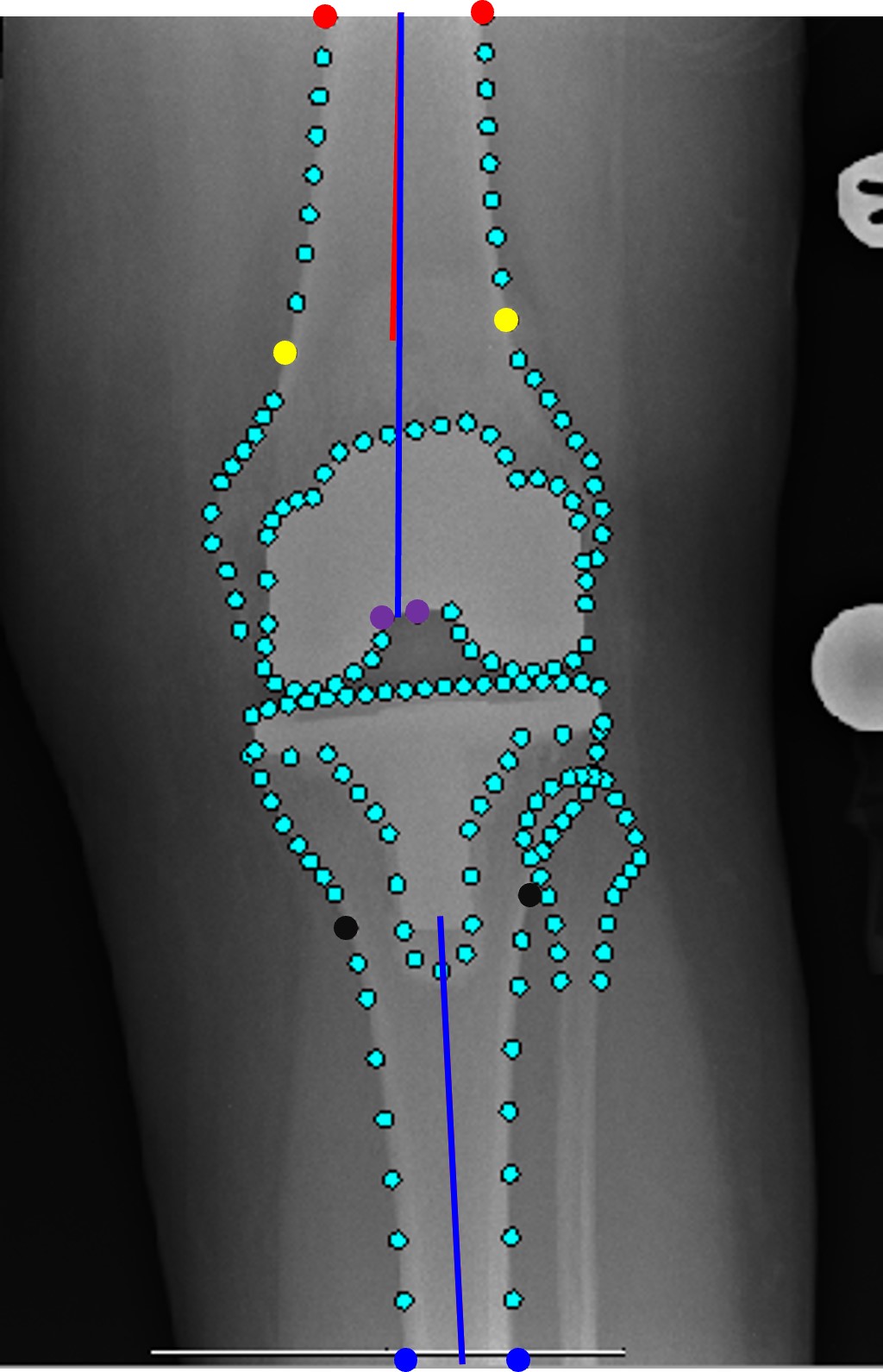}
    }
\caption{An illustration of the two calculation methods of the point-based KA measurements in (a) pre-operative and (b) post-operative images. In both (a) and (b), FTS fits a red center line to the femur and tibia by connecting two shaft center points (femur: the mid-points of the red and yellow point pairs; tibia: the mid-points of the black and blue point pairs), whereas FNTS fits a blue line to the femur by connecting a shaft center point (mid-point of the red point pair) to a femoral notch point (mid-point of the purple point pair), and to the tibia by connecting two shaft center points (mid-points of the black and blue point pairs). The blue lines may overlap the red lines.}
\label{fig:KA_egs}
\end{figure}

We investigated two calculation methods for the point-based measurements: one (\textbf{FTS}) using only femoral and tibial shaft points, and another (\textbf{FNTS}) incorporating femoral notch information with the femoral and tibial shaft points. The two calculation methods in pre-operative and post-operative knee radiographs are visualized in Fig.~\ref{fig:Preop_KA} and~\ref{fig:Postop_KA}, respectively.

\section{Experiments}

\subsection{Implementation Details}
Hourglass networks were trained using PyTorch 2.3.1 for deep learning-based pre-operative and post-operative knee radiograph analysis, with 220 and 120 epochs for global search and 800 and 600 epochs for local search, respectively. Global search models were trained on NVIDIA Tesla V100 GPUs, while local search models were trained on NVIDIA A100 GPUs. The network widths (initial numbers of channels) were 32 and 256 for global and local search, respectively. Wing loss~\cite{feng2018wing} was applied to emphasize small errors, and the model was optimized with Adam~\cite{kingma2014adam} using a learning rate of 0.0001. 

\subsection{Results}
\subsubsection{Landmark Localization}
We evaluated our landmark localization approach using relative point-to-point (rP2P) and relative point-to-curve (rP2C) distances. They were calculated to show the percentage of the reference length (tibial shaft width) defined by the distance between the two landmarks at the corners of the tibial plateau (see yellow points in Fig.~\ref{fig:workflow}). We had access to the pre-operative and post-operative RFRV-CLM models from~\cite{cullen2025an} to compare the landmark detection accuracy with our approach. The results are summarized in Table~\ref{landmarkresultap}. We found that our hourglass-based method could localize the knee landmarks more accurately and robustly than~\cite{cullen2025an} with lower rP2P and rP2C distances. 

\begin{table}[ht]
\centering
\caption{Quantitative comparison of pre-operative and post-operative landmark localization accuracy in AP knee radiographs between the proposed method and~\cite{cullen2025an}. The relative point-to-point (rP2P) and relative point-to-curve (rP2C) distances were calculated to show the percentage of the reference length (tibial shaft width) defined by the distance between the two landmarks at the corners of the tibial plateau (see yellow points in Fig.~\ref{fig:workflow}).}\label{landmarkresultap}
\begin{tabular}{|p{2.5cm}|p{2.5cm}|c|c|c|c|c|c|c|c|c|c|}
\hline
\multirow{2}{*}{\makebox[2.5cm][c]{Data}} & \multirow{2}{*}{\makebox[2.5cm][c]{Method}} & \multicolumn{3}{|c|}{rP2P} & \multicolumn{3}{|c|}{rP2C} \\
\cline{3-8}
 & & Mean & Median & 95$\%$ile & Mean & Median & 95$\%$ile \\
\hline
\multirow{2}{*}{\makebox[2.5cm][c]{Pre-operative}} & \centering RFRV-CLM~\cite{cullen2025an} & 4.1$\%$ & 3.4$\%$ & 9.5$\%$ & 1.1$\%$ & 0.6$\%$ & 2.1$\%$ \\
& \centering \textbf{This study} & \textbf{1.7$\%$} & \textbf{1.6$\%$} & \textbf{2.5$\%$} & \textbf{0.6$\%$} & \textbf{0.5$\%$} & \textbf{0.8$\%$} \\
\hline
\multirow{2}{*}{\makebox[2.5cm][c]{Post-operative}} & \centering RFRV-CLM~\cite{cullen2025an} & 3.6$\%$ & 2.5$\%$ & 11.7$\%$ & 1.3$\%$ & 0.6$\%$ & 9.0$\%$ \\
& \centering \textbf{This study} & \textbf{1.6$\%$} & \textbf{1.6$\%$} & \textbf{2.4$\%$} & \textbf{0.4$\%$} & \textbf{0.4$\%$} & \textbf{0.6$\%$} \\
\hline
\end{tabular}
\end{table}

\subsubsection{Alignment Measurement}
Intra-class corelation coefficient (ICC), mean absolute difference (MAD), and Bland-Altman analysis (BAA) were used to assess the agreement between the automated measurements generated from the automatically localized landmark positions and the two sets of ground truth measurements. Higher ICC and lower MAD or BAA bias indicate better performance. In Table~\ref{gtvsdl}, we summarize the results of our KA measurement experiments, and compare the results to those presented in~\cite{cullen2025an}.

\begin{table}[ht]
\centering
\caption{Agreement between automated and manual/clinical aTFA measurements calculated by \textbf{FTS} and \textbf{FNTS}. The best performances, compared with clinical measurements, are highlighted with *. (Pre-op: Pre-operative; Post-op: Post-operative; M: manual; A: automated; C: clinical; ICC: intra-class correlation coefficient; MAD: mean absolute difference; BAA: Bland-Altman analysis; n: number of individuals; CI: confidence interval; SD: standard deviation)}\label{gtvsdl}
\begin{tabular}{|c|c|l|c|c|c|c|c|c|}
\hline
\multirow{2}{*}{\makebox[1cm][c]{\shortstack{aTFA}}} & \multirow{2}{*}{\makebox[1.4cm][c]{Method}} & \multirow{2}{*}{\makebox[3.5cm][c]{Agreement}} & \multicolumn{2}{|c|}{ICC} & \multicolumn{2}{|c|}{MAD} & \multicolumn{2}{|c|}{BAA} \\
\cline{4-9}
& & & Value & CI 95\% & Value & SD & Bias & SD \\
\hline
\multirow{8}{*}{\makebox[1cm][c]{FTS}} & \multirow{4}{*}{\makebox[1.4cm][c]{\shortstack{RFRV- \\ CLM~\cite{cullen2025an}}}} & Pre-op M and A (n=376)  & 0.97 & (0.96, 0.97) & 1.0° & 1.3° & 0.2° & 1.6° \\
& & Post-op M and A (n=376) & \textbf{0.88} & (0.86, 0.90) & 0.9° & 1.1° & 0.2° & 1.4° \\
& & Pre-op C and A (n=50) & \textbf{0.95} & (0.91, 0.97) & 1.4° & 1.4° & \textbf{-0.8°} & 1.8° \\
& & Post-op C and A (n=50) & 0.78 & (0.67, 0.86) & 1.5° & 1.3° & 0.5° & 2.0° \\
\cline{2-9}
& \multirow{4}{*}{\makebox[1.4cm][c]{\shortstack{This \\ study}}} & Pre-op M and A (n=376)  & \textbf{0.99} & (0.99, 0.99) & \textbf{0.6°} & 0.7° & \textbf{0.1°} & 1.0° \\
& & Post-op M and A (n=376) & 0.83 & (0.80, 0.86) & \textbf{0.6°} & 1.6° & \textbf{0.0°} & 1.7° \\
& & Pre-op C and A (n=50) & \textbf{0.95} & (0.91, 0.98) & \textbf{1.3°} & 1.3° & \textbf{-0.8°} & 1.7° \\
& & Post-op C and A (n=50)* & \textbf{0.86} & (0.76, 0.92) & \textbf{1.0°} & 1.3° & \textbf{0.2°} & 1.6° \\
\hline
\multirow{8}{*}{\makebox[1cm][c]{FNTS}} & \multirow{4}{*}{\makebox[1.4cm][c]{\shortstack{RFRV- \\ CLM~\cite{cullen2025an}}}} & Pre-op M and A (n=376)  & 0.98 & (0.97, 0.98) & 0.8° & 1.1° & 0.2° & 1.4° \\
& & Post-op M and A (n=376) & 0.92 & (0.90, 0.93) & 0.8° & 0.8° & \textbf{0.1°} & 1.1° \\
& & Pre-op C and A (n=50) & \textbf{0.97} & (0.95, 0.98) & \textbf{1.2°} & 1.0° & \textbf{0.1°} & 1.6° \\
& & Post-op C and A (n=50) & 0.71 & (0.56, 0.81) & 1.7° & 1.5° & 0.8° & 2.2° \\
\cline{2-9}
& \multirow{4}{*}{\makebox[1.4cm][c]{\shortstack{This \\ study}}} & Pre-op M and A (n=376)  & \textbf{0.99} & (0.99, 0.99) & \textbf{0.6°} & 0.6° & \textbf{0.0°} & 0.8° \\
& & Post-op M and A (n=376) & \textbf{0.96} & (0.95, 0.97) & \textbf{0.5°} & 0.6° & \textbf{0.1°} & 0.7° \\
& & Pre-op C and A (n=50)* & \textbf{0.97} & (0.95, 0.98) & \textbf{1.2°} & 1.0° & \textbf{0.1°} & 1.6° \\
& & Post-op C and A (n=50) & \textbf{0.81} & (0.69, 0.89) & \textbf{1.2°} & 1.4° & \textbf{0.5°} & 1.7° \\
\hline
\end{tabular}
\end{table}

\paragraph{FTS}
When analyzing the agreement between manual/clinical and automated measurements (Table~\ref{gtvsdl}), the pre-operative ICC values showed excellent agreement (>0.9), whereas the post-operative ICC values showed good agreement (0.75-0.9). The MAD values indicated minimal deviations ($\mathord{\sim}1$°) pre-operatively and post-operatively. The BAA showed no bias (<1°). Our method outperformed~\cite{cullen2025an} in most cases, except for the post-operative ICC value between manual and automated measurements.

\paragraph{FNTS}
When analyzing the agreement between manual/clinical and automated measurements (Table~\ref{gtvsdl}), only the post-operative ICC value between clinical and automated measurements showed good agreement (0.75-0.9), while other ICC values showed excellent agreement (>0.9). The MAD values indicated minimal deviations ($\mathord{\sim}1$°) pre-operatively and post-operatively. The BAA showed no bias (<1°). Our method outperformed~\cite{cullen2025an} with a higher ICC value, as well as lower MAD value and similar BAA bias. In most cases, FNTS demonstrated better agreement than FTS except for the post-operative agreement between clinical and automated measurements.

\section{Discussion and Conclusion}

We developed an automated system to localize knee anatomical landmarks and measure KA. To our knowledge, this is the first deep learning-based system to localize over 100 knee landmarks, fully outlining the knee joint while integrating KA measurements on AP knee radiographs. Our results on pre-operative and post-operative images from 376 TKR patients show that our hourglass-based system achieves consistently improved performance in localization accuracy compared with~\cite{cullen2025an}. 

The system demonstrates excellent accuracy and reliability in measuring varus/valgus KA. Our method achieves better performance than~\cite{cullen2025an} except for the post-operative ICC value between manual and automated measurements when calculating the aTFA with FTS. Post-operative agreement is lower than pre-operative agreement in terms of ICC, especially when comparing clinical and automated measurements. The automated measurements show a higher agreement with the manual measurements compared to the clinical measurements, possibly because of additional considerations in clinical practice like limb deformities instead of only using point position-based information. When calculating the aTFA, incorporating the femoral notch information can improve the overall reliability except when assessing the post-operative agreement between clinical and manual measurements.

As clinical measurements of only a single expert were used as reference in this study, in the future additional clinical measurements should be added to analyze the clinical variation in the future. A limitation of this study is that the system has not been tested for generalizability on another dataset. It would be of interest to use our trained models to generate KA measurements on an independent dataset. 

KA is strongly associated with TKR outcomes. For example, both varus and valgus post-operative malalignment were found to be associated with a higher incidence of revision surgery in several studies~\cite{fang2009coronal,ritter2011effect,ritter2013preoperative}. Future work will explore the relationship between KA measurements and TKR outcomes, aiming to predict surgical outcomes such as chronic pain or revision surgery in advance based on KA measurements in knee radiographs. In addition, the automatically localized landmark positions enable more complex analysis of knee joint shape and alignment (e.g. via Statistical Shape Models~\cite{cootes1995active}), beyond of what can be currently captured by a set of geometric measurements. This opens up opportunities for better utility of the information contained in AP knee radiographs, enabling more efficient and appropriate treatment decisions.

%\section*{Acknowledgments}

\clearpage

\begin{comment}  %% removed for anonymized MICCAI 2025 submission.
    
    % The following acknowledgement and disclaimer sections should be removed for the double-blind review process.  
    % If and when your paper is accepted, reinsert the acknowledgement and the disclaimer clause in your final camera-ready version.

\begin{credits}
\subsubsection{\ackname} A bold run-in heading in small font size at the end of the paper is
used for general acknowledgments, for example: This study was funded
by X (grant number Y).

\subsubsection{\discintname}
It is now necessary to declare any competing interests or to specifically
state that the authors have no competing interests. Please place the
statement with a bold run-in heading in small font size beneath the
(optional) acknowledgments\footnote{If EquinOCS, our proceedings submission
system, is used, then the disclaimer can be provided directly in the system.},
for example: The authors have no competing interests to declare that are
relevant to the content of this article. Or: Author A has received research
grants from Company W. Author B has received a speaker honorarium from
Company X and owns stock in Company Y. Author C is a member of committee Z.
\end{credits}

\end{comment}

\begin{thebibliography}{8}
\bibitem{jin2020incidence}
Jin, Z., Wang, D., Zhang, H., Liang, J., Feng, X., Zhao, J. and Sun, L.: Incidence trend of five common musculoskeletal disorders from 1990 to 2017 at the global, regional and national level: results from the global burden of disease study 2017. Annals of the rheumatic diseases \textbf{79}(8), 1014--1022 (2020)

\bibitem{ozden2025what}
Özden, V.E., Osman, W.S., Morii, T., Pastor, J.C.M., Abdelaal, A.M. and Younis, A.S.: What percentage of patients are dissatisfied post-primary total hip and total knee arthroplasty?. The Journal of Arthroplasty \textbf{40}(2), S55--S56 (2025)

\bibitem{defrance2023are}
DeFrance, M.J. and Scuderi, G.R.: Are 20$\%$ of patients actually dissatisfied following total knee arthroplasty? A systematic review of the literature. The Journal of Arthroplasty \textbf{38}(3), 594--599 (2023)

\bibitem{ritter2011effect}
Ritter, M.A., Davis, K.E., Meding, J.B., Pierson, J.L., Berend, M.E. and Malinzak, R.A.: The effect of alignment and BMI on failure of total knee replacement. JBJS \textbf{93}(17), 1588--1596 (2011)

\bibitem{ritter2013preoperative}
Ritter, M.A., Davis, K.E., Davis, P., Farris, A., Malinzak, R.A., Berend, M.E. and Meding, J.B.: Preoperative malalignment increases risk of failure after total knee arthroplasty. JBJS \textbf{95}(2), 126--131 (2013)

\bibitem{lindner2014robust}
Lindner, C., Bromiley, P.A., Ionita, M.C. and Cootes, T.F.: Robust and accurate shape model matching using random forest regression-voting. IEEE transactions on pattern analysis and machine intelligence \textbf{37}(9), 1862--1874 (2014)

\bibitem{lindner2013accurate}
Lindner, C., Thiagarajah, S., Wilkinson, J.M., arcOGEN Consortium, Wallis, G.A. and Cootes, T.F.: Accurate bone segmentation in 2D radiographs using fully automatic shape model matching based on regression-voting. In Medical Image Computing and Computer-Assisted Intervention–MICCAI 2013: 16th International Conference, Nagoya, Japan, September 22-26, 2013, Proceedings, Part II 16, pp. 181--189. Springer Berlin Heidelberg (2013). 

\bibitem{tiulpin2019kneel}
Tiulpin, A., Melekhov, I. and Saarakkala, S.: KNEEL: Knee anatomical landmark localization using hourglass networks. In Proceedings of the IEEE/CVF International Conference on Computer Vision Workshops, pp. 0--0 (2019). 

\bibitem{oktay2018attention}
Oktay, O., Schlemper, J., Folgoc, L.L., Lee, M., Heinrich, M., Misawa, K., Mori, K., McDonagh, S., Hammerla, N.Y., Kainz, B. and Glocker, B.: Attention u-net: Learning where to look for the pancreas. arXiv preprint arXiv:1804.03999. (2018)

\bibitem{newell2016stacked}
Newell, A., Yang, K., and Deng, J.: Stacked hourglass networks for human pose estimation. arXiv preprint arXiv:1603.06937. (2016)

\bibitem{feng2018wing}
Feng, Z.H., Kittler, J., Awais, M., Huber, P. and Wu, X.J.: Wing loss for robust facial landmark localisation with convolutional neural networks. In Proceedings of the IEEE/CVF conference on computer vision and pattern recognition, pp. 2235--2245 (2018). 

\bibitem{kingma2014adam}
Kingma, D.P. and Ba, J.:  Adam: A method for stochastic optimization.  arXiv preprint arXiv:1412.6980. (2014)

\bibitem{cullen2025an}
Cullen, D., Thompson, P., Johnson, D. and Lindner, C.: An AI-based system for fully automated knee alignment assessment in standard knee AP radiographs. The Knee \textbf{54}, 99--110 (2025)

\bibitem{ronneberger2015unet}
Ronneberger, O., Fischer, P. and Brox, T.: U-Net: Convolutional Networks for Biomedical Image Segmentation. arXiv preprint arXiv:1505.04597. (2015)

\bibitem{davison2019landmark}
Davison, A.K., Lindner, C., Perry, D.C., Luo, W., Medical Student Annotation Collaborative and Cootes, T.F.: Landmark localisation in radiographs using weighted heatmap displacement voting. In Computational Methods and Clinical Applications in Musculoskeletal Imaging: 6th International Workshop, MSKI 2018, Held in Conjunction with MICCAI 2018, Granada, Spain, September 16, 2018, Revised Selected Papers 6, pp. 73–85. Springer International Publishing (2019).

\bibitem{payer2019integrating}
Payer, C., Štern, D., Bischof, H. and Urschler, M.: Integrating spatial configuration into heatmap regression based CNNs for landmark localization. Medical Image Analysis \textbf{54}, 207--219 (2019)

\bibitem{ye2020development}
Ye, Q., Shen, Q., Yang, W., Huang, S., Jiang, Z., He, L. and Gong, X.: Development of automatic measurement for patellar height based on deep learning and knee radiographs. European Radiology \textbf{30}, 4974--4984 (2020)

\bibitem{nam2023key}
Nam, H.S., Park, S.H., Ho, J.P.Y., Park, S.Y., Cho, J.H. and Lee, Y.S.: Key-point detection algorithm of deep learning can predict lower limb alignment with simple knee radiographs. Journal of Clinical Medicine \textbf{12}(4), 1455 (2023)

\bibitem{fang2009coronal}
Fang, D.M., Ritter, M.A. and Davis, K.E.: Coronal alignment in total knee arthroplasty: just how important is it?. The Journal of arthroplasty \textbf{24}(6), 39--43 (2009)

\bibitem{cootes1995active}
Cootes, T.F., Taylor, C.J., Cooper, D.H. and Graham, J.: Active shape models-their training and application. Computer vision and image understanding \textbf{61}(1), 38--59 (1995)

\end{thebibliography}
%

\end{document}